\pdfoutput=1

\documentclass[11pt]{article}

\usepackage[preprint]{acl}

\usepackage{times}
\usepackage{latexsym}

\usepackage[symbol]{footmisc}

\usepackage[T1]{fontenc}

\usepackage[utf8]{inputenc}

\usepackage{microtype}

\usepackage{inconsolata}

\usepackage{graphicx}

\title{Investigating Human-Aligned Large Language Model Uncertainty}

\author{
   Kyle Moore\footnote[2] \\ \\
   Vanderbilt University \\
   \texttt{kyle.a.moore@vanderbilt.edu} 
   \\ \And
  Jesse Roberts\footnote[2] \\ \\
  Tennessee Tech University \\
  \texttt{jtroberts@tntech.edu} 
   \\ \AND
   Daryl Watson \\
  Tennessee Tech University \\
   \\ \And
   Pamela Wisniewski \\
   Vanderbilt University \\
}

\usepackage[most]{tcolorbox}
\tcbset{
    mymsg/.style={
        enhanced,
        colframe=white, frame hidden,
        boxrule=0pt,
        arc=18pt,
        outer arc=18pt,
        boxsep=2pt, left=8pt, right=8pt, top=6pt,
        colback=#1!25,
        fuzzy shadow={1mm}{-0.7mm}{0mm}{0.1mm}{black!40!white},
        width=0.8\linewidth,
        before skip=-0.5mm, 
    },
    expertstyle/.style={mymsg=green, enlarge left by=10mm},
    llmstyle/.style={mymsg=blue, enlarge right by=10mm}
}

\begin{document}
\maketitle
\begin{abstract}
    Recent work has sought to quantify large language model uncertainty to facilitate model control and modulate user trust. Previous works focus on measures of uncertainty that are theoretically grounded or reflect the average overt behavior of the model. In this work, we investigate a variety of uncertainty measures, in order to identify measures that correlate with human group-level uncertainty. We find that Bayesian measures and a variation on entropy measures, top-k entropy, tend to agree with human behavior as a function of model size. We find that some strong measures decrease in human-similarity with model size, but, by multiple linear regression, we find that combining multiple uncertainty measures provide comparable human-alignment with reduced size-dependency.
    
\end{abstract}

\section{Introduction}
    \footnotetext[2]{Equal Contribution}
\label{sec:intro}
There has been a rise as of late in research interest in appropriate and effective measures of uncertainty in large language models (LLMs). Accurate measures of LLM uncertainty  are essential for calibrating and improving trust in the models. In particular, uncertainty measures provide important context for assessing the safety of models as components in safety-critical systems and mitigating overtrust by end users. In both cases, uncertainty measures are necessary due to the common phenomenon of hallucination, a pernicious behavior that has dominated discussions concerning LLMs both in the research community and the broader public.

Prior works, which we survey extensively in section \ref{sec:prior-work}, have focused largely on uncertainty in terms of model task efficacy. It is comparatively rare that existing works discuss uncertainty in a manner grounded in human behavior. Our work seeks to fill this gap by investigating whether and to what extent measures of LLM uncertainty match the uncertainty signature seen among human groups. Many have investigated human-like behavioral phenomena in LLMs. Importantly, this includes behaviors that are well accepted in psychological research as uncertainty-born, including typicality effects \citep{misra2021language, roberts2024using, vemuri2024well} and fan effects \citep{roberts2024large}. To our knowledge, this is the first work that attempts to directly compare human and LLM uncertainty on a paired task.

Uncertainty quantification in LLMs has typically sought to characterize and calibrate uncertainty according to an objective standard of correctness. While this is useful in many contexts, human intuition of uncertainty is often disparate from the actual values. For example, while a model would be justified in claiming it is mostly certain if it has a certainty level of $60\%$, human users, based on typical usage of the phrase, may expect a certainty level of $\sim90\%$. This disparity may affect human propensity to place too much trust on model outputs. If there exists an uncertainty measure that is well calibrated to human uncertainty, referred to here as \textit{aligned}, reporting such a measure could yield benefits for human-computer interaction and human-AI collaboration.

In this work, we make the following contributions:
\begin{enumerate}
    \item We directly compare human uncertainty against a diverse variety of LLM uncertainty measures, lending insight those that are aligned in general and those that are only aligned in large models.
    \item We propose uncertainty measures that have not, to our knowledge, been investigated in LLMs, including nucleus size, top-k entropy, and choice entropy, as well as derivative uncertainty measures.
    \item We find evidence that mixtures of uncertainty measures can provide comparable human alignment to the top isolated uncertainty measures without the associated size dependency.
\end{enumerate}

\section{Background}
\label{sec:background}

This paper investigates uncertainty quantification (UQ) in large language models. Here we provide some background on LLMs and uncertainty, while a survey of existing UQ methods will be reserved for section \ref{sec:prior-work}.

\subsection{Large Language Models}
Current LLM architectures are based broadly on the transformer introduced by \citet{vaswani2017attention}. While the original transformer architecture utilized a traditional encoder-decoder structure, current models are overwhelmingly encoder-only or decoder-only models. Encoder-only models, typified by BERT \citep{devlin2019bert}, are largely relegated to masked language modeling tasks where one or more tokens at any position in a given context are masked or otherwise unavailable. The model's task is to predict the most appropriate tokens to insert in those masked positions.

Most well-known LLMs are decoder-only models \cite{roberts2024powerful}. These are causal models which predict, given the current context, the most appropriate next token to add to the end of the current context. They are trained and used autoregressively by iteratively predicting a new token, adding that new token to the original context, and repeating the process until either a maximum number of tokens are generated or a special stop token is predicted. 

We investigate uncertainty exclusively in decoder-only language models. Specifically, we examine the LLaMa 3.1 \citep{grattafiori2024llama}, LLaMa 3.2 \citep{llama32}, Mistral 0.1, and Mistral 0.3 \citep{jiang2023mistral7b} models. For each of these, we investigate both the base completion and the instruction-finetuned (instruct) versions. The instruct models are finetuned to respond in a conversational manner and follow directional instructions in addition to standard next-word prediction. Both varieties of a given model see heavy usage in practice, with base models being used for specialized backend applications and instruct models being used in many user-facing applications.

Another important distinction in LLMs is between black box and white box models. Black box models, typified by OpenAI's ChatGPT \citep{chatgpt} and Anthropic's Claude \citep{claude} models, do not provide access to a model's internal weights. Black box models often will only provide an output sequence and may not provide even the token probability distribution for each generation step. Because of these limitations, black box models, despite being the primary type of model with which users interact, are difficult to study in meaningful detail. 

White box models may be split into open-weight and open-source models. These are differentiated primarily in that open-weight models provide only model weights and architecture, while open-source releases additionally provide all code and data needed to train the model from scratch. In both cases, white box models allow users to inspect the internal structure, weights, and activation of models at inference time. This makes white box models more conducive and appropriate for research into model behaviors. Therefore, all models discussed in this work are open-weight white box models.

\subsection{Uncertainty}

Uncertainty refers to a broad collection of processes and behaviors in which decisions must be made without access to perfect information. Uncertainty quantification (UQ) is a branch of research within machine learning (ML) that attempts to quantify the amount of uncertainty in a system in order to inform downstream tasks, characterize task accuracy, etc. Uncertainty in deep ML models has been extensively studied \cite{gawlikowskiSurveyDeep2023}, but the complexity of LLM behavior presents unique UQ challenges. We discuss the existing UQ methods for LLMs in section \ref{sec:prior-work}.

Most extant research decomposes uncertainty based on its source into epistemic and aleatoric uncertainty. Epistemic uncertainty refers to uncertainty that results from missing or noisy information intrinsic to the model as a result of the specifics of the training process. Aleatoric uncertainty instead describes uncertainty sourced from incomplete information about the task being performed, such as uncertainty due to imperfect sensors. In the context of LLMs, this is usually framed as a distinction between knowledge left out of the training corpus (epistemic) and knowledge left out of the current context (aleatoric) \citep{beigi2024rethinking}. This work does not make this distinction, instead seeking to quantify total uncertainty regardless of source.

In line with the safety and hallucination focus in much of the extant uncertainty research, many works seek to measure or improve a model's uncertainty calibration. Calibration refers to how closely a model's quantified uncertainty matches with its likelihood to output a correct answer in a given context. The goal is typically that the model's measured uncertainty should match the amount of relevant knowledge to which the model has access. In the context of human-like uncertainty behaviors, calibration may be an irrelevant or inappropriate goal. Humans are capable of expressing both complete certainty and complete uncertainty regardless of correctness. We sidestep this issue entirely by focusing on contexts in which no available answer is more correct than any other. In particular, as described in section \ref{sec:methods-data}, we compare model uncertainty against human survey response data.

\section{Prior Work}
\label{sec:prior-work}

Uncertainty quantification in LLMs is a rapidly growing area of research that involves a wide variety of methods and downstream goals. Of particular prevalence are works that seek to use uncertainty measures to facilitate model calibration to task accuracy \cite{lin2022teaching, mielke2022reducing, band2024linguistic, tian2023just, chaudhry2024finetuning, steyvers2025large, jiang2021can} and to effectively guide human trust in the model's outputs \cite{steyvers2025large, belem2024perceptions}. Calibration refers to the problem of inducing models to, given a chosen uncertainty measure, yield uncertainty levels in a given task that correlate with the model's accuracy on that task. Although the benefits of calibration are clear for general task completion, it is not clear that calibration is desirable for models that are intended to show human-like behavior.
While uncertainty and task accuracy may be correlated in humans across a variety of tasks, they are not necessarily linked.
Humans are capable of displaying low uncertainty with low accuracy as well as high uncertainty despite consistently high accuracy \cite{dunning2011dunning}. To our knowledge, this work is the first to investigate LLM uncertainty \textit{aligned} to human uncertainty in contexts where there is no correct answer.

Related work also exists that compares human and LLM response patterns, though not uncertainty, on non-factual datasets. These works have primarily made such comparisons for the purposes of identifying model biases \citep{tjuatja2024llms}, eliciting model opinions \citep{dominguez2024questioning,santurkar2023whose,durmus2023towards}, or simulating survey responses for downstream research \citep{argyle2023out,huang2025uncertainty}. Of these, the most similar work to ours is \citet{huang2025uncertainty}, which used uncertainty-aware comparisons to generate human-like responses. Our work differs in that we compare numerous uncertainty measures with human responses to identify which measures direct and derived measures are best aligned to human uncertainty.

\subsection{LLM Uncertainty Measurement}
\label{sec:unc-meas}
Methods for measuring uncertainty in LLMs fall into at least five identifiable methodological categories: self-reporting, consistency, logit-based, entropy-based, and ensemble-based. Each of these has been explored to varying degrees and in a variety of contexts.

Self-reported uncertainty measures are the most common UQ method in the literature. These measures rely on the models to provide a measure of their own uncertainty. These methods typically fall into two strategies. The more common strategy is the tendency of a model to include in its output a phrase indicating certainty (PIC). The PIC may be explicit (typically expressed as a percentage)  \citep{chaudhry2024finetuning, shrivastava2023llamas, tian2023just, xiong2023can, belem2024perceptions} or implicit (typically expressed as one of a collection of phrases; e.g. "I think it's ..." or "I'm sure it's ...") \citep{zhou2023navigating, mielke2022reducing, band2024linguistic, lin2022teaching, tang2024evaluation}. Alternative to PIC-based strategies are self-assessment methods, which query the model in two or more stages. The first stage obtains an answer from the model to the initial query. For some variations, the second stage evaluates whether the initial answer is (in)correct \citep{lin2022teaching,kadavath2022language,chen2024quantifying}. Other variations instead generate multiple alternative answers in the second stage and, in a third stage, query how many of the second-stage answers factually agree with the initial answer \citep{manakul2023selfcheckgpt, zhang2024luq}.

Consistency-based UQ measures are common for black-box LLMs because it requires no access to model internals. In these measures, the model is queried repeatedly with the same context. The model's certainty is determined by the percentage of responses that contain the most commonly provided answer \citep{kaur2024addressing,khan2024consistency,lin2023generating}. These are similar to the latter self-reported methods, but they typically do not use multiple generation stages and often use manual inspection to determine answer overlap. Given a large number of inferences, consistency-based methods are statistically guaranteed to converge to logit-based methods given identical prompting. This is because LLMs are deterministic apart from the separate method which chooses the next token from the probability density output by the LLM.

Logit-based UQ measures directly inspect the model's logit outputs or some transformation over those logits, most typically softmax to obtain a probability density function (pdf) over the available tokens. The simplest methods use the relative probability of the most probable output as the level of certainty \citep{shrivastava2023llamas, jiang2021can, steyvers2025large, tian2023just}. Some work has sought to use surrogate models to approximate logit-based methods for black-box models \citep{shrivastava2023llamas}.

Entropy-based methods measure uncertainty in model outputs using Shannon entropy, defined as $-\sum_{x\in X}p(x)\log p(x)$, where $X$ is a collection of candidate output tokens. Entropy-based methods have seen limited usage in LLM uncertainty estimation despite their simplicity and usage in other machine learning domains. While entropy is defined primarily for single-token generation, sentence-level entropy can be obtained most simply by taking the average or maximum entropy over all generation steps \citep{huang2025look}. \citet{kadavath2022language} briefly introduces a variant on sentence-level entropy based on the accumulation of token-wise entropy. \citet{duan2024shifting} extends this further by weighting each token by a calculated relevance score.

Ensemble methods are born out of a stream of research into Bayesian uncertainty estimation. Traditional methods for general neural networks involve aggregating the predictions of multiple independently trained models \citep{lakshminarayanan2017simple, xiao2021hallucination, malinin2020uncertainty}, but this is nonviable for current LLMs due to high training time and costs. Methods that approximate this process have been developed using Low-Rank Adapters (LoRA) \citep{wang2023lora} or Monte Carlo dropout to generate ensembles at inference time \citep{roberts2024using, gal2016dropout, fomicheva2020unsupervised}. Other works have attempted to approximate ensembles for uncertainty estimation using test-time augmentation of contexts \citep{hou2024decomposing}.


\section{Methods}
\label{sec:methods}
In this section, we will describe our experimental methodology. This will begin with a discussion of how we constructed our dataset, followed by a detailed description of each of the uncertainty measures tested, and finishing with the finer details of our experiment design decisions and prompting methodology.

\subsection{Dataset}
\label{sec:methods-data}
Our data was manually collected from the Pew Research Center's American Trends Panel (ATP) Datasets \citep{pew}. We opted for human surveys on non-factual questions so that we could isolate the uncertainty correlations without potential confounds from differing knowledge access between the human subjects and the LLMs. In this way, we can compare the LLM uncertainty directly with human disagreement on questions without a designated correct answer.

We collected data from the the 20 most recent waves of survey data at time of writing, encompassing waves 113 through 132 and covering a human data collection time period of August 1, 2022 through August 6, 2023. We filtered questions from the various waves according to the following criteria. We removed free-response and write-in questions that do not have a predetermined set of allowed answers to facilitate cloze comparison. We omitted questions to which the human responses could change drastically over short time-frames, such as COVID-19 surveys or political approval ratings. We removed personal experience and history questions that a model would not have the proper background data to support. Finally, we removed questions that targeted responses from specific subpopulations (e.g. Asian-American men under 30), which may show markedly different response distribution from untargeted populations. This included for omission some survey questions that provided response data for the total population, but split the results into demographic subgroups and did not provide subgroup sizes to permit reconstructing the total response distribution.

The results of this process were 38 questions pulled from 8 survey waves, representing over 500000 human response data points. The minimum number of human responses to a question was 5079, with a maximum of 30861. The response distributions were diverse, including questions with near unanimous agreement and near complete disagreement. Question and answer texts were minimally edited from the versions given to human survey participants for agreement with the base query format.

\subsection{Experiment Design}
\label{sec:methods-exp-des}

\begin{figure}[h!]
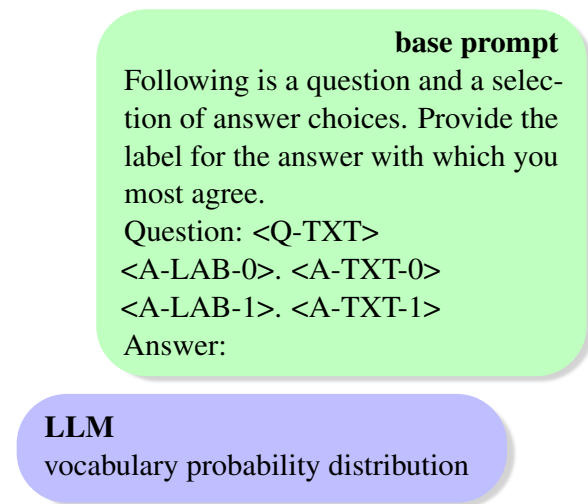

    \centering
    \resizebox{\linewidth}{!}{ 
    \begin{tcolorbox}[expertstyle]
        \hfill \textbf{base prompt} \\
        Following is a question and a selection of answer choices. Provide the label for the answer with which you most agree. \\
        Question: <Q-TXT> \\
        \ <A-LAB-0>. <A-TXT-0> \\
        \ <A-LAB-1>. <A-TXT-1> \\
        Answer: 
    \end{tcolorbox}
    } 

    \vspace{2mm} 

    \resizebox{\linewidth}{!}{ 
    \begin{tcolorbox}[llmstyle]
        \textbf{LLM} \\
        vocabulary probability distribution

    \end{tcolorbox}
    } 
    \caption{Prompt to measure presence/absence belief.}
    \label{fig:Prompt}
\vskip-0.5em

\end{figure}

All LLM experiments used a shared base query format defined as shown in figure \ref{fig:Prompt}. For each question in the dataset, we insert the question text in place of \textit{<Q-TXT>}. For each associated answer, we insert the $i$-th answer's content in place of \textit{<A-TXT-i>} and replace \textit{<A-LAB-i>} with successive alphabetic labels such that \textit{<A-LAB-0>} = ``A'', \textit{<A-LAB-1>} = ``B'', and so on.

The result is a base query for each question in the dataset. We pass the base queries to the LLM to obtain a probability distribution over the entire vocabulary. For measures that require the model's chosen answer, we obtain this using cloze testing over the answer choice label tokens. The chosen answer choice is taken to be the answer choice whose label is more probable than any other choice label. For ensemble methods, a cloze test is performed for each ensemble variant and the model's chosen answer is taken to be the answer label most often chosen using cloze testing across the entire ensemble. This results in a chosen answer and a probability distribution over all tokens in the vocabulary. The next section details how these are used to calculate each of our chosen uncertainty measures.

\subsection{Uncertainty Measures}
\label{sec:methods-unc}
In our experiments, we used eight candidate measures of uncertainty. Those candidate uncertainty measures are labeled as self-reported (SR), response frequency (RF), Nucleus Size (NS), vocabulary entropy (VE), choice entropy (CE), top-k entropy (KE), population variance (PV), and population self-reported (PS). With this set, each of the broad uncertainty measure categories is represented: self-report (SR, PS), consistency-based (RF), logit-based (RF, TP, PV), entropy-based (VE, CE, KE), and ensemble-based (PV, PS). In this section, we describe how each of these measures were obtained.

\subsubsection{Self-reported Measures}
Our approach to SR uncertainty uses a method inspired by \citet{chen2024quantifying}. They measure uncertainty in a two-stage process. First, the model is queried using the base query to obtain a chosen answer. This answer is then appended to the context followed by a secondary query that evaluates the probability that the model evaluates the initial answer as correct, incorrect, or a variety of other evaluator phrases. These probabilities are aggregated into a final uncertainty measure. To account for the non-factual nature of the questions in our dataset, we borrow from \cite{roberts2024large2} by using the evaluator phrases ``best'' and ``worst'' to gauge relative preference across available choices. Higher certainty should yield lower probability for ``worst'' and higher probability for ``best''.

\subsubsection{Consistency and Logit Measures}
RF uncertainty is measured as the probability of the label tokens associated with each answer choice. For single-token generation tasks, the probability of each label token multiplied by a number of trials is the expected number of repeated inferences in which that token is generated by the model, meaning that this measure covers both logit-based uncertainty and an idealized variation on consistency-based uncertainty. Higher probability of the target token is taken to indicate higher certainty.

\begin{figure*}[ht]
    \centering
    \includegraphics[width=\linewidth]{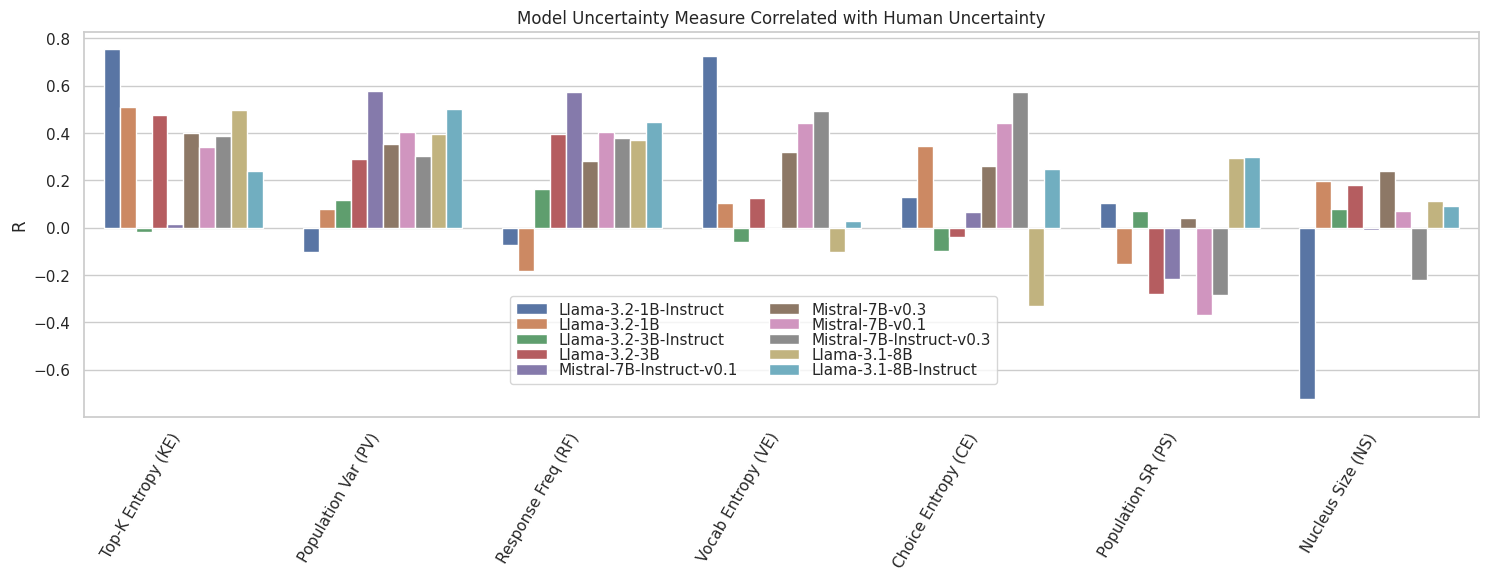} 
    \caption{Correlation between uncertainty in human response data and LLM uncertainty across all uncertainty measures. Measures are ordered by mean correlation across models.}
    \label{fig:sim}
\end{figure*}

\begin{figure}[ht]
    \centering
    \includegraphics[width=\linewidth]{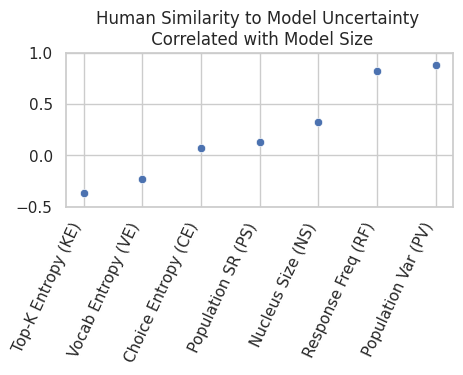} 
    \caption{Correlation between measure human-similarity and model size across all models. Measures are ordered by correlation with model size.}
    \label{fig:size}
\end{figure}

The nucleus (commonly known as the top-p) of a probability distribution is the set of the most probable tokens given the base query such that their summed probability is less than some specified probability threshold \citep{holtzman2019curious}. NS is the number of tokens contained in the nucleus. This measure is equivalent to the credible interval, which is the Bayesian analog to confidence intervals. In line with this, we use a nucleus threshold of $0.95$. Intuitively, this measure represents the number of tokens that the model considers to be reasonable options, with a larger NS (more reasonable candidates) indicating higher uncertainty. To our knowledge, this is the first work to use NS as an indicator of uncertainty.

\subsubsection{Entropy Measures}

The entropy methods - VE, CE, and KE - all use Shannon entropy over the vocabulary probability distribution given the base query. For VE, this is computed over the entire probability distribution $VE=\sum_{v_i\in V}P(v_i|q_b)\log P(v_i|q_b)$ where $V$ is the set of all vocabulary tokens and $q_b$ is the base query. CE and KE are variations on entropy that restrict the tokens considered to a subset of the vocabulary. KE computes entropy over only the $k$ most probable tokens $V_k$, such that $KE=\sum_{v_i\in V_k}P(v_i|q_b)\log P(v_i|q_b)$. In our experiments, we chose to use a fixed $k=10$. CE similarly restricts the vocabulary to a subset, $V_c$, that are pre-selected based on the task. In our experiments, $V_c$ contains exactly the labels associated with the answer choices provided in the base query ($V_c = {A,B,C}$ for questions with three available answers.) In all cases, these provide a measure of the imbalance in the probability distribution. High entropy indicates relatively equal probability across all candidate and thus low certainty. To our knowledge, KE and CE are novel uncertainty measures in the context of LLMs.

\subsubsection{Ensemble Measures}
In our experiments, we create ensembles of stochastic variations on each model using Monte Carlo dropout \citep{roberts2024using}. We generate a new ensemble of $N=30$ variants per base model for each question in the dataset. PS uncertainty is obtained by following the same procedure defined for SR for each ensemble variant. PV uncertainty is obtained by taking the standard deviation across ensemble variants of the probability of each candidate token. This can be expressed as $PV=\sigma_{e\in E}(P_e(v_a|q_b))$, where $E$ is the set of ensemble variants, $P_e$ is the probability function associated with variant $e$, and $v_a$ is the token associated with answer $a$. We measure the certainty for each of the choice labels ($a\in V_c$).
\begin{figure*}[h!]
    \centering
    \includegraphics[width=\linewidth]{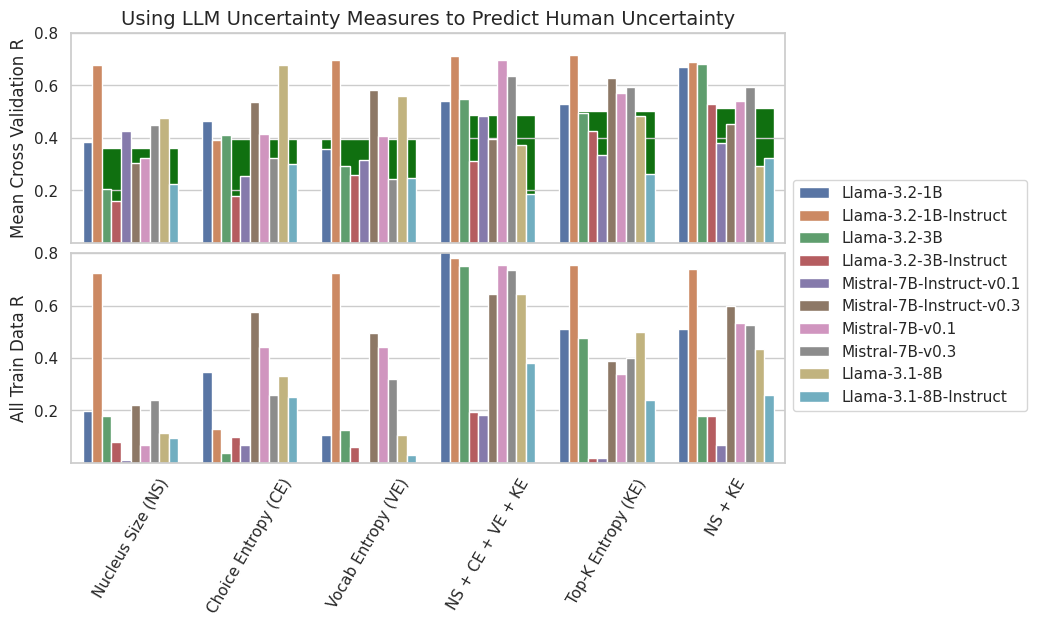} 
    \caption{Accuracy of linear regression models trained on LLM measuered uncertainty and predicting human uncertainty. Top: 3-fold cross validation, dark green background bar indicates the mean correlation across models. Models ordered by mean correlation. Bottom: Results when model trained and tested on the entire dataset.}
    \label{fig:IPF}
\end{figure*}
\section{Results}

In this section, we present the results of our experiments described in section \ref{sec:methods}. We split our results and analysis into two phases. In the first phase, we evaluate to what degree the human and LLM uncertainty measures are similar in terms of correlation across questions. In phase two, we evaluate the ability of the various measures to predict human uncertainty.

\subsection{Human-LLM Uncertainty Similarity}

Our initial analysis of the various uncertainty measures is shown in Figure \ref{fig:sim}. For every question, the uncertainty in the human response data is taken as the entropy over the set of available answer choices, $\sum_{a\in A}P_h(a|q)\log P_h(a|q)$ where $A$ is the set of answer choice labels and $P_h(a|q)$ is the percent of human subjects who selected answer $a$ on question $q$. Figure \ref{fig:sim} shows the correlation across all questions between the human uncertainty measure and each of the candidate LLM uncertainty measures, ordered by mean correlation across models. To obtain the correlation between human uncertainty and the LLM population, the standard deviation of the logit output across the population for each options is compared to the relative frequency of each answer among the human population. In line with existing literature \cite{hinkle2003applied}, we set a significance threshold of $|r|\geq0.3$.

Only measures KE, PV, and RF achieve significance for the majority of models. All three measures show a large effect of model size on human-similarity, though KE notably correlates negatively such that smaller models align better than larger models. NS and CE show significance across most models without a strong dependence on model size. The relationship between model size and human-LLM uncertainty similarity is shown in more detail in figure \ref{fig:size}, in which we correlated the correlation scores from figure \ref{fig:sim} with the size of the model. NS and VE stand out here as being significantly correlated across sizes despite low human-similarity.

\subsection{Human Uncertainty Prediction}

To further investigate the connection between human uncertainty and our uncertainty measures, we attempted to learn a set of simple linear regression models. We also repeat this process using 3-fold cross validation to assess generalizability. Ideally, a measure which was aligned with human uncertainty would correlate well on the entire dataset when trained and correlate well with validation data. Otherwise, the model may correlate with some but not all the data or have untenable model variance, respectively.

For this phase, we omitted PV, PS, and RF because they are computed on a per-choice basis and cannot be readily adapted to linear regression. One linear regression model is learned per model-measure pair, with the human uncertainty on each question as the dependent variable and uncertainty measure value on the associated questions as the independent variables. Further, we train two additional linear regressions that predict human uncertainty using a linear combination of multiple measures. One linear regression uses all uncertainty measures except RF, PV, and PS, while the other is trained using only the best and worst performing measures in phase one: KE and NS. This pair of measures also has potentially relevant features including that they (1) have similar but opposite correlation with model size and (2) fall cleanly into vocabulary uncertainty (NS) and choice uncertainty (KE) measures. 

The results of this process are shown in figure \ref{fig:IPF}, ordered left to right in ascending order by mean accuracy across models. The relative performance of each measure matches the relative human-similarity across measures as in figure \ref{fig:sim}. Importantly, the combination of all non per-choice measures results in a model that generalizes well, $r \approx 0.5$, and has significant correlation, $r > 0.6$, with the entire dataset. This is the only measure which manages a balance of both. This may be because the measures, taken as a group, balance increasing and decreasing correlation with model size as shown in figure \ref{fig:size}.


\section{Discussion}
\label{sec:disc}
While we see positive results on the similarity between LLM uncertainty and human uncertainty for a variety of measures, we see across all analyses that the most effective measure for achieving human-similarity in uncertainty response is KE. RF and PV also show evidence of a strong dependence on model size for increased human-similarity, suggesting that they might be effective measures for larger ($>8$ billion parameters) models, but further study is warranted to verify this relationship holds. PV presents an issue here, as ensemble methods scale poorly with increased model size. KE similarly appears to scale badly with increased model size despite being the most human-similar measure at the model sizes studied here. The results in phase two of analysis suggests that mixtures of measures may be able to mitigate their respective weaknesses and provide a stronger overall measure.

\section{Future Work}

This work leaves open many avenues of future research. The most obvious research path is extending this work to compare other, more sophisticated, uncertainty measures from the existing literature. Similarly, future work should consider iterations on the simplistic KE and RF methods to achieve higher human-similarity, such as dynamic selection of k. In both cases, future work should seek to extend this work to a wider variety of contexts, as the current work is limited to only instantaneous responses with a limited set of available completions. Future work may extend this work to contexts that allow or require unconstrained generation.



\section*{Limitations}
\label{sec:limits}

In this section, we identify some of the limitations in our study that may affect generalization of the results. All methods used herein rely on the model under study being capable of providing the probability distribution over the output vocabulary. Access to this distribution is not provided by most black-box models. As such, we cannot replicate these results on black-box models and the human-like uncertainty measures identified cannot be transferred. Some of the methods, particularly the ensemble methods, also require access to the model internals and so cannot be used with black-box models. Even within the domain of white-box models, our experiments were limited to only LLaMa and Mistral model families. We cannot make guarantees on transferability to other model families. 

Our experiment design also makes use of an unstated assumption that the uncertainty behavior of groups of human respondents is directly comparable to the uncertainty behavior of an individual LLM. It is unclear from our work whether these results are transferable to comparisons between the uncertainty of an LLM with the uncertainty of an individual human subject. However, our results show that both individual and group LLM uncertainty (PV and KE in particular) correlates with human group behavior. This provides some confidence that our results may be transferable.

Finally, our experiment design utilized cloze testing on a set of pre-defined tokens for ease of comparison. We cannot guarantee without future research that uncertainty correlations hold for unconstrained generation, either in terms of full vocabulary generation or for intended responses.


\bibliography{references}




\end{document}